\documentclass[letterpaper]{article}
\usepackage{aaai}
\usepackage{times}
\usepackage{helvet}
\usepackage{courier}
\usepackage{graphicx}
\usepackage{amssymb}
\usepackage[numbers]{natbib}
\let\svthefootnote\thefootnote
 
\begin{document}
%
\title{A Hybrid State Space-Attention Architecture for Forecasting the Trajectories of Battery Degradation}
\author{Shrey Patel, Bhargav Srinivas Rao, Samet Ayhan\\
College of Computational Mathematical Natural Sciences\\
University of Maryland - College Park\\
College Park, MD, USA\\
}
\maketitle
\let\thefootnote\svthefootnote
\begin{abstract}
\begin{quote}
Early-life battery degradation trajectory forecasting aims to predict a battery cell's full state-of-health (SOH) trajectory from its initial charge–discharge cycles, potentially reducing cell qualification and evaluation time from months to days. Despite recent advances, accurate forecasting remains challenging for emerging battery chemistries, data-scarce domains, and cells exhibiting degradation behaviors beyond those observed during training. Existing state-of-the-art approaches increasingly rely on model scaling, large language model-derived aging-condition embeddings, and extensive fold-specific hyperparameter tuning, raising concerns about computational efficiency, reproducibility, and generalization. This work introduces DecayFormer, a compact hybrid state-space and attention architecture designed for early-life degradation trajectory forecasting. DecayFormer combines cycle-level selective recurrence, using a single scalar state per channel, with an attention mechanism queried by the recurrent state. It further incorporates state-of-charge (SOC) interval statistics and a master curve head that jointly estimates a scalar lifetime and a monotonically non-increasing SOH profile parameterized by normalized lifetime. We evaluate DecayFormer on the BatteryLife benchmark using all 22 aging-condition-exclusive folds, with a single run per fold and one fixed configuration across all folds, without fold-specific tuning. DecayFormer achieves mean test MAPE values of 2.25\% on Li-ion, 2.76\% on CALB, 0.95\% on Na-ion, and 7.19\% on Zn-ion. These results demonstrate the potential of compact sequence models to deliver accurate early-life degradation forecasts across diverse battery chemistries and aging conditions, without relying on extensive model scaling or per-fold optimization. 
\end{quote}
\end{abstract}

\section{Introduction}
A battery's degradation governs how it can be deployed, what warranty it can carry, and how quickly a new chemistry can be developed, and measuring that degradation takes months of cycling. Early degradation trajectory forecasting shortens this by predicting a cell's full SOH trajectory from its first cycles \cite{BatteryMFormer}. The task is hardest where emerging chemistries have few labeled cells \cite{BatteryLife}, and the early signs of degradation are confined to particular SOC intervals of the voltage-current curve \cite{BatteryMFormer}. Recent methods address these difficulties mainly through scale and side information.

Forecasting a trajectory is considerably harder than forecasting its end point, because capacity fade is rarely linear. Many cells degrade slowly and almost linearly for most of their life and then enter a regime of accelerated fade, the knee, whose onset varies widely between nominally identical cells and between operating conditions \cite{AttiaKnees,FerminCueto}. The underlying mechanisms, chiefly the loss of lithium inventory and the loss of active electrode material, leave distinct signatures in the voltage curve long before they are visible in capacity \cite{Birkl}, and their relative weight differs between cells, protocols and chemistries. A trajectory forecaster must therefore predict both when a cell degrades and how its degradation unfolds, and it must do so for thousands of cycles from an observation window of at most one hundred.

Two further difficulties shape the design of such a forecaster. First, the forecast horizon is long relative to the information available, which favours compact representations that summarise the observed cycles without overfitting them; structured state space models \cite{S4,Mamba} provide such summaries at a cost linear in sequence length, and hybrids that combine recurrence with attention \cite{Jamba,Griffin} have shown that the two mechanisms are complementary. Second, test cells drawn from unseen aging conditions may age faster or slower than any cell seen in training, so the forecaster must extrapolate in lifetime. The physical sciences meet an analogous problem through superposition: responses measured under different conditions collapse onto a single master curve once time is rescaled by a condition dependent shift factor \cite{WLF}. Embedding such structure in a learned model, rather than relying on scale alone, is one route to combining physical insight with data driven prediction \cite{Aykol}.

We propose DecayFormer\footnote{Code and reproducibility notebooks: \texttt{https://github.com/ShreyPatel1311/DecayFormer}. Trained weights of all 22 fold models: \texttt{https://huggingface.co/Godseye1311/DecayFormer}.}, a compact hybrid of state space recurrence and attention architectures. It applies a selective recurrence over cycle tokens whose per channel state is a single scalar, paired with an attention whose query comes from that recurrent state, and with SOC-interval statistics adapted from BatteryMFormer \cite{BatteryMFormer}. Its central idea is a master curve head, which predicts lifetime as a scalar and a monotonically non-increasing SOH profile over fraction of life, so that under a lifetime shift only the scalar has to extrapolate. DecayFormer has 0.79M parameters, about 6\% of BatteryMFormer's, and uses a single configuration on every fold, with no per fold hyperparameter search, no test protocol metadata and no language model.

Our contributions include:
\begin{itemize}
\item DecayFormer, a trajectory forecaster whose only inputs are the charge and discharge curves of each early cycle (voltage, current and capacity, resampled along SOC) together with three measured per cycle scalars, built around a state queried attention block that supports long horizon forecasting from a single scalar state per channel.
\item A master curve head that separates lifetime from degradation shape, targeting robustness when lifetimes differ between training and test cells.
\end{itemize}

\section{Related Works}

Most battery prognostics work predicts a single number, that is, the cycle life of a cell, or its remaining useful life at a given point \cite{Severson,BatteryLife}. Forecasting the full SOH trajectory from early cycles is the more demanding task, and it can only be trained and compared fairly on a benchmark that supplies full trajectories, spans several chemistries, and fixes splits in which test cells come from aging conditions unseen in training \cite{BatteryMFormer}.

Data driven approaches to battery prognostics have progressed from probabilistic regression to deep sequence models. Gaussian process regression forecasts state of health with calibrated uncertainty from capacity measurements alone \cite{Richardson}, and recurrent networks with long short-term memory model the long range dependence of capacity fade for remaining useful life prediction \cite{ZhangLSTM}, a task to which transformer architectures have also been applied \cite{ChenTransformer}. Closest to the present task, Li et al.\ \cite{LiOneShot} predict the entire degradation trajectory in a single pass from approximately one hundred early cycles, including the knee and the end of life. Because the knee governs much of the error of a trajectory forecast, it has also been studied directly: Ferm\'{\i}n-Cueto et al.\ \cite{FerminCueto} define the knee point and knee onset from capacity data and predict both from the first fifty cycles, and Attia et al.\ \cite{AttiaKnees} review the mechanisms that produce knees. These methods are, however, developed and evaluated on single chemistry Li-ion datasets with their own splits, which motivates the benchmark adopted here.

A growing body of work forecasts the whole trajectory from early cycles. Rieger et al.\ \cite{Rieger} predict the complete degradation curve autoregressively from the initial cycles, with ensemble uncertainty estimates and an explainability analysis. Chen et al.\ \cite{ChenJoint} jointly predict cycle life, with a multiple output Gaussian process, and the capacity trajectory, with a prompt learning network conditioned on the predicted end point; the separation of a lifetime scalar from a degradation profile in DecayFormer follows a related decomposition. Because complete trajectories are scarce, several methods augment the training data: Lin et al.\ \cite{LinSynthetic} train a transferred convolutional network on synthetic trajectories, Meng et al.\ \cite{MengFewShot} pretrain on pseudo trajectories generated from a single cell and adapt to new scenarios with few shots, and Jin et al.\ \cite{JinCGAN} generate synthetic data with a conditional generative adversarial network for a transformer whose architecture and hyperparameters are configured by evolutionary search. Du et al.\ \cite{DuSiamese} predict lifespan with a Siamese convolutional network from partial voltage and current data of the first 30 cycles and use it as a prior for trajectory prediction, and Yang et al.\ \cite{YangTemporal} use temporal attention to generalise across operating conditions, with hyperparameters selected by Bayesian optimisation. At larger scale, BatteryGPT \cite{BatteryGPT} autoregressively generates charging data across the battery lifetime with a generative pretrained transformer and estimates SOH from the generated data. These methods are developed on Li-ion cells, and several depend on synthetic data, pretraining or hyperparameter optimisation; DecayFormer uses none of these and a single configuration across 22 folds and four chemistries.

Public battery aging data has grown steadily, but most datasets were built for other purposes. The NASA Prognostics Center of Excellence data \cite{NASA} and the CALCE data \cite{CALCE} cover a small number of Li-ion cells under a few conditions and are mostly used for capacity estimation and remaining useful life prediction. The Oxford degradation data \cite{Oxford} similarly covers only a handful of cells. Severson et al.\ \cite{Severson} released 124 commercial LFP/graphite cells cycled to failure under fast charging protocols, and established early cycle life prediction as a task; the follow up by Attia et al.\ \cite{Attia} added cells from closed-loop protocol optimisation. Later releases added further Li-ion cells and conditions, but each dataset uses its own format, its own preprocessing and its own splits, and nearly all of them are a single chemistry. Results reported on them are rarely comparable with one another. Two efforts address this fragmentation. BatteryML \cite{BatteryML} provides an open platform that unifies several public Li-ion datasets under one preprocessing and evaluation pipeline, but its benchmark centres on cycle life prediction. BatteryLife \cite{BatteryLife} integrates a larger set of datasets in one format and extends coverage beyond commercial Li-ion cells to Na-ion and Zn-ion chemistries and to the CALB industrial cells. Among the datasets examined, BatteryLife is the only one that provides per cycle SOH trajectories across multiple chemistries with published aging condition exclusive folds. BatteryMFormer \cite{BatteryMFormer} builds its trajectory forecasting benchmark on this data, and that benchmark is adopted for this research.

Severson et al.\ \cite{Severson} showed that the variance of the difference between discharge capacity voltage curves of two early cycles, $\Delta Q(V)$, predicts cycle life well with a regularised linear model, which motivates per interval curve statistics. General time series forecasters such as PatchTST \cite{PatchTST}, DLinear \cite{DLinear} and iTransformer \cite{iTransformer} have also been applied, and BatteryLife \cite{BatteryLife} and BatteryMFormer \cite{BatteryMFormer} benchmark them alongside battery specific models, including the CyclePatch variants CPMLP and CPTransformer introduced by BatteryLife \cite{BatteryLife} and the intra-cycle and inter cycle enhanced model IC2ML \cite{IC2ML}. BatteryMFormer \cite{BatteryMFormer} is the current reference model for trajectory forecasting on this benchmark. It learns at several levels of the data, including a view over SOC intervals across cycles that its own ablation identifies as its largest single contributor, and it conditions on a language model embedding of each cell's test protocol text. It uses 12.9 to 14.9M parameters depending on the fold, with a hyperparameter search per fold. A different route to sequence modelling is the selective, input dependent linear recurrence of Mamba \cite{Mamba}, which Mamba2 \cite{Mamba2} shows to be equivalent to causal attention under a structured decay mask (state space duality); PhyMamba \cite{phyMamba} applies a physics modulated Mamba to battery health prognostics.

Two further lines of work inform the design of DecayFormer. On the side of physical structure, degradation diagnostics extract the loss of lithium inventory and of active material from open circuit voltage and incremental capacity curves \cite{Birkl}, establishing that the voltage curve carries mechanistic information, and Aykol et al.\ \cite{Aykol} survey architectures that combine physics based and machine learning models to obtain lifetime predictions that remain consistent with physical constraints. Master curves obtained by time temperature superposition \cite{WLF} provide the template for separating a rate from a shape. On the side of sequence architecture, structured state space models \cite{S4} established linear recurrences as competitive long sequence models, and Mamba \cite{Mamba} made them selective; hybrid language models such as Jamba \cite{Jamba}, which interleaves Transformer and Mamba layers, and Griffin \cite{Griffin}, which mixes gated linear recurrences with local attention, show that recurrence and attention are complementary at scale. DecayFormer applies this hybrid principle at the level of charge-discharge cycles, in a regime of few labelled cells rather than abundant text.

\begin{figure*}[t]
  \centering
  \includegraphics[width=1.5\columnwidth]{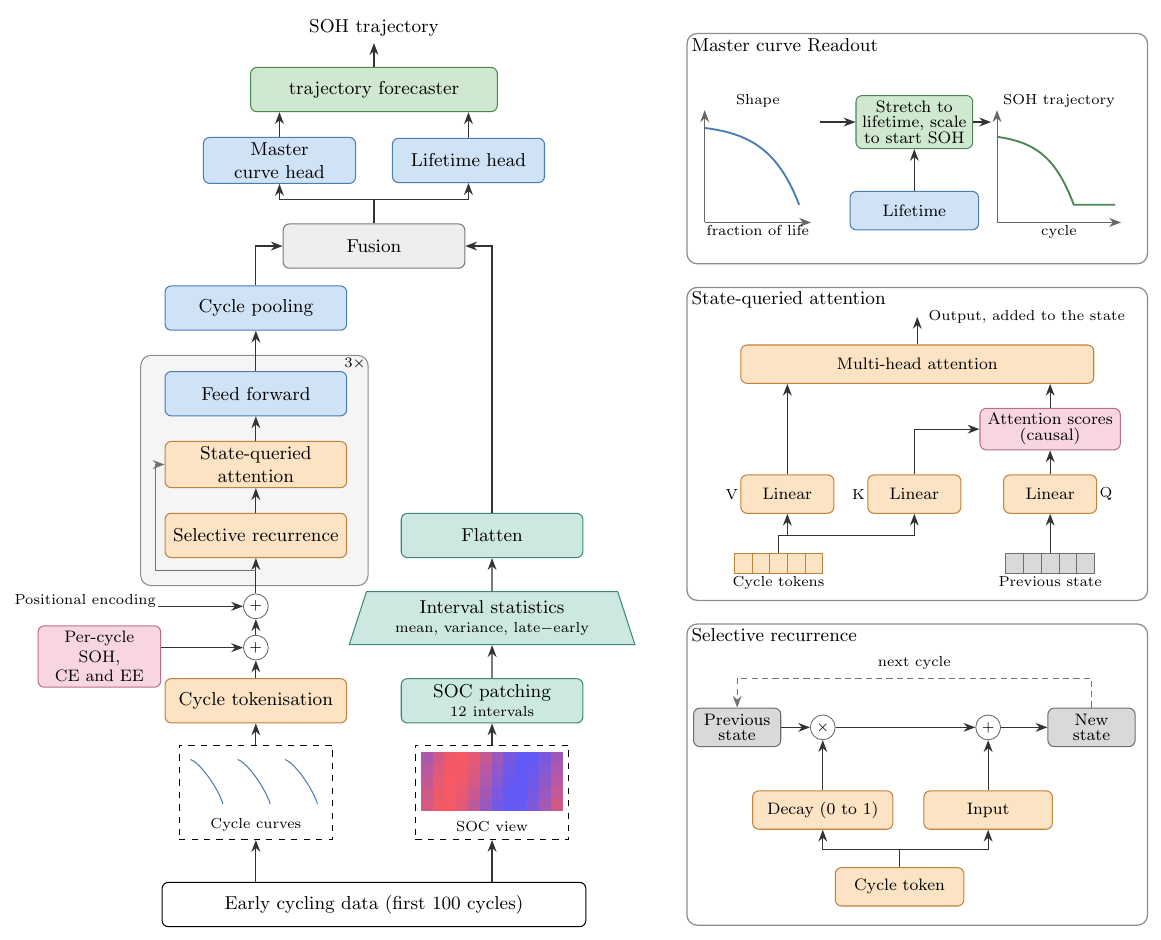}
  \caption{Cycle tokens pass through $N$ layers of selective recurrence and state queried attention; a parallel SOC localised view summarises each SOC interval across cycles; the master curve head outputs a non increasing SOH trajectory.}
  \label{fig:architecture}
\end{figure*}

\section{Methodology}
DecayFormer takes the first $S\le100$ cycles of a cell and predicts its SOH at every subsequent cycle up to a horizon of $T=5000$; training and evaluation are restricted to cycles the cell attained. The model, shown in Figure~\ref{fig:architecture}, comprises 791,529 parameters.

\subsection{Tokenization}
Each of the $S$ observed cycles is resampled to $L=300$ points along the SOC axis. A cycle is therefore represented as an $L\times C$ matrix with $C=4$, comprising three curve channels, voltage, current and capacity over the charge and discharge segments, together with the SOC curve. Stacking the cycles yields $x\in\mathbb{R}^{S\times L\times C}$, accompanied by binary mask indicating observed cycles. Three measured per cycle scalars are additionally provided: coulombic efficiency, energy efficiency, and the measured SOH of the observed cycles. 

The $L\times C$ matrix of each cycle is flattened to a vector of dimension $LC=1200$ and mapped to a token $z_s\in\mathbb{R}^{d}$ with $d=96$ by a linear projection followed by two residual feedforward blocks of width $2d$, yielding one token per cycle as in CyclePatch \cite{BatteryLife}. The three measured scalars are added to the token, and a sinusoidal positional encoding over the cycle index is applied. The encoder consists of $N=3$ identical blocks over the cycle axis, each residual and with normalization applied before its sublayers, with 4 attention heads and dropout 0.1.

\begin{table*}[t]
  \centering
  \caption{Test MAPE (\%, lower is better) on the four BatteryLife domains under BatteryMFormer's aging condition exclusive fold protocol. All models were executed in BatteryMFormer's released pipeline at one run per fold, at the fold's own seed. Values are reported as mean${\pm}$std across the folds of each domain; the dispersion therefore characterises variation between folds rather than between runs. The final row states the number of folds on which DecayFormer attains the lowest MAPE of the four models.}
  \small
  \label{tab:final_result}
  \begin{tabular}{lcccc}
    \hline
    Model & Li-ion (3) & CALB (4) & Na-ion (12) & Zn-ion (3) \\
    \hline
    BatteryMFormer & $2.421_{\pm0.160}$ & $2.809_{\pm2.052}$ & $1.099_{\pm0.402}$ & $\mathbf{5.907}_{\pm1.348}$ \\
    IC2ML          & $2.692_{\pm0.265}$ & $\mathbf{2.361}_{\pm1.297}$ & $1.944_{\pm0.789}$ & $6.172_{\pm0.615}$ \\
    CPMLP          & $2.565_{\pm0.090}$ & $3.434_{\pm1.906}$ & $1.366_{\pm0.479}$ & $7.526_{\pm1.549}$ \\
    \textbf{DecayFormer} & $\mathbf{2.252}_{\pm0.111}$ & $2.763_{\pm2.095}$ & $\mathbf{0.950}_{\pm0.303}$ & $7.185_{\pm0.385}$ \\
    \hline
    DecayFormer best of four & 3/3 & 2/4 & 8/12 & 0/3 \\
    \hline
  \end{tabular}
\end{table*}

\subsection{State Attention Encoder Block}
Each block applies two causal operations over the cycle axis in sequence. The two differ in the origin of their attention weights. $\odot$ denotes the elementwise product, $\sigma$ the logistic sigmoid, and $\tilde z_s$ the layer normalized token; $W_\Delta$, $W_b$, $W_g$, $W_q$, $W_C$ and $W_D$ are learned linear maps.

\emph{Selective Recurrence} Each channel propagates a scalar state,
\begin{equation}
h_s = a_s \odot h_{s-1} + b_s,\qquad a_s=\exp\!\big(-\Delta_s\odot e^{\alpha}\big)\in(0,1),
\end{equation}
where $\Delta_s=\mathrm{softplus}(W_\Delta \tilde z_s)$ is a strictly positive
step size, $b_s=(W_b \tilde z_s)\odot\Delta_s$, and $\alpha$ is a learned per
channel parameter. Since $\Delta_s$ and $b_s$ depend on the input, the retention
of past information is determined per cycle rather than fixed, following the
selectivity mechanism of Mamba \cite{Mamba}. The block propagates the gated
residual $z_s+\sigma(W_g \tilde z_s)\odot(h_s+\gamma\odot \tilde z_s)$, in which
$\gamma$ is a learned per channel skip. Unobserved cycles are assigned $a_s=1$
and $b_s=0$, leaving the state unmodified.

The recurrence is linear in $h$ and therefore admits the closed form
\begin{equation}
h_s=\sum_{j\le s} \Lambda_{sj}\, b_j,\qquad \Lambda_{sj}=\prod_{k=j+1}^{s} a_k,
\end{equation}
which constitutes a causal attention operation whose weight matrix $\Lambda$ is a learned decay mask dependent on the input, rather than a softmax over products of queries and keys. This is the state space duality of Mamba-2 \cite{Mamba2}, the recurrence is evaluated as the prefix scan of Hillis and Steele \cite{HillisSteele} with $O(\log S)$ sequential depth, and explicit materialization of $\Lambda$ computes the same function to floating point tolerance.

\emph{State queried attention.} A multi head softmax attention subsequently
attends over the cycle tokens, with the query derived from the recurrent state
rather than from the current token, and specifically from the state prior to the
incorporation of cycle $s$:
\begin{equation}
h'_s = h_s + \mathrm{Attn}\big(q=W_q h_{s-1},\; z_{1:s}\big).
\end{equation}
Querying with $h_s$ would permit the query to attend over content it already
encodes. Keys and values are obtained from the cycle tokens $z_{1:s}$ by further
learned linear maps, and the attention output is incorporated into the state
rather than only into the readout, so that retrieved information propagates to
all subsequent cycles. The block emits $y_s = W_C h'_s + W_D z_s$, followed by a
feedforward layer with normalization applied beforehand. Following the final
block, the cycle tokens are pooled over observed cycles by their mean and log
variance; the second moment is included because the strongest hand crafted
predictor reported on this data is itself a variance \cite{Severson}.

\subsection{SOC branch}
Adapted from BatteryMFormer \cite{BatteryMFormer}, this branch operates on the raw curves rather than cycle tokens and transposes two axes, producing one token per SOC interval summarized across cycles. A convolution layer with kernel and stride $P=25$ partitions each cycle into $M=L/P=12$ interval tokens, each constructed exclusively from points within the SOC windows. Interval $m$ is then characterized by 3 masked statistics over the observed cycles namely, mean, log variance, and a contrast $\delta_m$ between late and early cycles, whose early and late quartiles are derived from cumulative observation mask so that prefixes of differing length yield comparable contrasts. The quantity $\delta_m$ constitutes a learned per interval generalization of $Q_{\mathrm{late}}(V)-Q_{\mathrm{early}}(V)$, the difference between late and early discharge capacity at a given voltage $V$ \cite{Severson}. The $M$ interval tokens are flattened rather than pooled, preserving interval identity and projected to $d$ dimensions.

\subsection{Master curve head}
The pooled cycle branch representation and the projected SOC branch representation are concatenated to form $\psi\in\mathbb{R}^{3d}$, which is supplied to 2 heads of identical structure. First head consists of a sequence of layer normalization, a linear map to $d$ dimensions, a GELU nonlinearity and a linear output map. The formulation corresponds to a time lifetime superposition, in which one head determines the rate at which a cell's degradation proceeds and the other determines the curve against which that rate is evaluated.

The master curve head emits $K=\lceil T/\tau\rceil=200$ values $\kappa_j$ at knots of stride $\tau=25$ cycles, which are transformed into a monotonically non-increasing curve of remaining health, not of degradation, on a normalized life axis $u\in[0,1]$:
\begin{equation}
s_k = 1-\frac{1}{K}\sum_{j\le k}\mathrm{softplus}(\kappa_j).
\end{equation}
Here $s(u)$ is the fraction of remaining health, equal to 1 at the start of the forecast; the corresponding degradation, $1-s(u)$, is cumulative and non-decreasing. The head therefore constrains SOH to be non-increasing, equivalently degradation to be non-decreasing. Normalization by $K$ causes the output bias to govern total capacity fade across the horizon rather than fade per knot. Capacity loss is cumulative, so emitting nonnegative decrements and integrating them renders the predicted SOH non-increasing by construction.

The lifetime head emits a single scalar $\ell$, the shift factor, obtained as
$\ell=\mathrm{clamp}(\exp(\cdot),10,10^5)$ with its final bias initialized at
$\log 500$, so that initialization corresponds to a plausible cycle life. The trajectory is then given by $\hat y_t = s\big(\mathrm{clamp}(t/\ell,0,1)\big)$
with linear interpolation between knots: $\ell$ supplies the sampling coordinates and $s$ the corresponding values, and the two outputs are not concatenated. The resulting curve is finally scaled by the mean measured SOH of the first five observed cycles, so that a cell commencing below nominal capacity, the Na-ion median being 0.898, is anchored at its observed initial level and $s$ is interpreted as the fraction of remaining health. Under a shift in cycle life, only $\ell$ is required to extrapolate, and the master curve transfers unchanged.

\begin{table}[t]
  \centering
  \caption{Ablation on the side inputs supplied to DecayFormer, under the folds and protocol of Table~\ref{tab:final_result}. The centre column reproduces DecayFormer as reported there. In the left column the measured side inputs are withheld, leaving only the cycling curves and the observed SOH. In the right column BatteryMFormer's Qwen3 embedding of the test protocol text is additionally supplied. MAPE (\%), reported as mean${\pm}$std across folds.}\small
  \label{tab:ablation}
  \setlength{\tabcolsep}{3pt}
  \begin{tabular}{lccc}
    \hline
    Domain & w/o side inputs & DecayFormer & w/ LLM embedding \\
    \hline
    Li-ion & $2.395_{\pm0.166}$ & $\mathbf{2.252}_{\pm0.111}$ & $2.358_{\pm0.137}$ \\
    CALB   & $2.687_{\pm2.134}$ & $2.763_{\pm2.095}$ & $\mathbf{2.408}_{\pm2.017}$ \\
    Na-ion & $\mathbf{0.947}_{\pm0.447}$ & $0.950_{\pm0.303}$ & $1.040_{\pm0.377}$ \\
    Zn-ion & $\mathbf{7.182}_{\pm0.739}$ & $7.185_{\pm0.385}$ & $7.409_{\pm0.533}$ \\
    \hline
  \end{tabular}
\end{table}
\begin{figure*}[t]
  \centering
  \includegraphics[width=1.6\columnwidth]{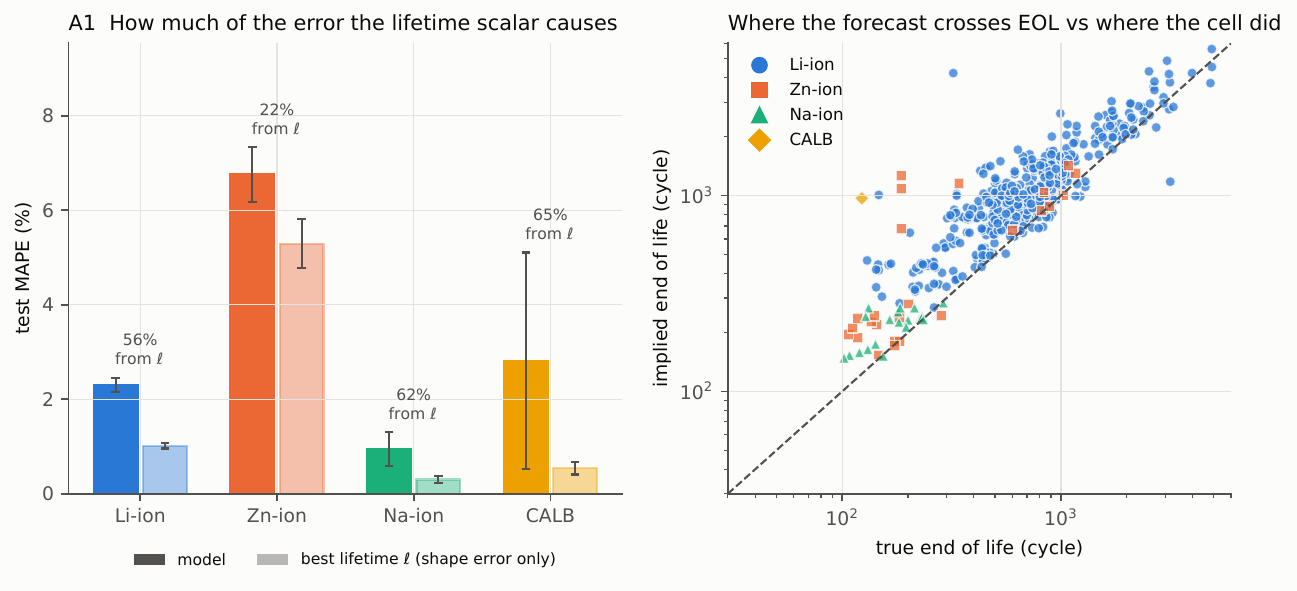}
  \caption{Error attributable to the lifetime scalar, over the test cells of all 22 folds (trained models, full precision evaluation). Left: test MAPE of DecayFormer (dark) and with $\ell$ re-chosen per sample while $s$ is held fixed (light), mean$\pm$std across folds; annotations give the share of error removed by correcting $\ell$ alone. Right: cycle at which the forecast crosses end of life against the true end of life, for cells whose data reach it (Spearman rank correlation \cite{Spearman} $\rho=0.84$ Li-ion, $0.78$ Zn-ion, $0.59$ Na-ion); dashed line marks equality.}
  \label{fig:oracle}
\end{figure*}

\section{Evaluation}
\subsection{Setup}
The model is evaluated on the benchmark that BatteryMFormer \cite{BatteryMFormer} builds on BatteryLife \cite{BatteryLife}, using its 22 aging condition exclusive folds unchanged: 3 on Li-ion, 4 on CALB, 12 on Na-ion and 3 on Zn-ion. Each fold assigns whole aging conditions, rather than individual cells, to training or to test: every cell of a test condition is held out, and no cell cycled under that condition is seen in training. The test cells of every fold therefore belong to aging conditions unseen in training, and each fold measures transfer to a new condition. Each model receives the first 100 cycles of a cell and forecasts its SOH trajectory, and the error is the mean absolute percentage error (MAPE) between the forecast and the measured SOH over the cycles the cell attained. DecayFormer uses the same training configuration and the same stopping criterion, early stopping on validation MAPE, on every fold, with no per fold tuning; each model is trained once per fold at that fold's own seed and evaluated on the fold's test cells. Each of the four values reported per model is an average across folds within one domain, the mean of the per fold test MAPE over that domain's folds, with the standard deviation across the same folds; no value is averaged across domains. We also report the number of folds on which DecayFormer attains the lowest MAPE. DecayFormer is compared with BatteryMFormer, the reference model on this benchmark, with IC2ML \cite{IC2ML}, and with CPMLP \cite{BatteryLife}, each run with its released configuration.

\emph{Computational cost.} Training DecayFormer on one fold, with early stopping, takes between approximately 3 and 52 minutes on a single 24 GB GPU (RTX 3090, RTX A5000 or RTX 4090): 3 to 4 minutes per CALB fold, 4 to 23 minutes per Na-ion fold, 7 to 36 minutes per Zn-ion fold and approximately 52 minutes per Li-ion fold. At the training batch of 128 in mixed precision, a training step allocates a peak of 1.2 GB of GPU memory, including the optimizer state. At inference, forecasting the full trajectory of one cell takes 18 ms at batch size one on an entry level NVIDIA T500 laptop GPU, and 1.2 ms per cell when 256 cells are forecast together, with a peak GPU memory allocation of 15 MB and 0.4 GB respectively; on the laptop's CPU alone, one cell takes 32 ms.

\subsection{Results}
Table~\ref{tab:final_result} reports DecayFormer alongside BatteryMFormer, IC2ML and CPMLP across all 22 folds. All four models were executed within BatteryMFormer's released pipeline, so that the comparison isolates the model class: the data, the target generation, the training loop, the early stopping criterion and the evaluation code are common to all of them. DecayFormer comprises 0.79M parameters, and employs a single configuration on every fold without per fold hyperparameter search.

DecayFormer attains the lowest mean MAPE of the four models on two domains, ranks second on a third and is outperformed on the fourth. The domains differ considerably in the number of folds available and in the dispersion across them, and these differences govern the strength of the conclusion that each supports; the domains are accordingly treated separately below. Na-ion comprises 12 of the 22 folds. DecayFormer attains 0.950, against 1.099 for BatteryMFormer, 1.366 for CPMLP and 1.944 for IC2ML, and achieves the lowest MAPE of the four models on 8 of 12 folds. Two folds, cond405 and cond411, are fully censored, in that their recorded trajectories terminate near SOH 1.0, so that no model is required to predict appreciable degradation upon them; both are nevertheless retained, as the protocol prescribes.

DecayFormer attains 2.252 and achieves the lowest MAPE of the four models on all three folds of Li-ion domain, ahead of BatteryMFormer at 2.421, CPMLP at 2.565 and IC2ML at 2.692. The fold dispersion, between 0.09 and 0.27, is small relative to the range of 0.44 separating the best and worst model. The four models span 2.361 to 3.434 on CALB domain, with IC2ML lowest, DecayFormer second at 2.763 and BatteryMFormer third at 2.809. A property of the release further restricts the inferences available from this domain: the public CALB data terminate at 99 to 100 cycles per cell, which coincides with the input length, so that the forecast window is short and the task correspondingly less demanding than on the remaining three domains. No claim is consequently advanced on CALB.

DecayFormer is outperformed on Zn-ion domain, attaining 7.185 against 5.907 for BatteryMFormer and 6.172 for IC2ML, and exhibiting a higher MAPE than both on all three folds; its error is lower only than that of CPMLP, at 7.526. The deficit relative to BatteryMFormer is 1.28 MAPE, which is comparable in magnitude to the fold standard deviation of 1.35, but it is consistent in sign across every fold and is therefore attributable to the model rather than to variation between folds. Zn-ion is additionally the smallest and most demanding domain in the benchmark, comprising 140 coin cells distributed over three folds with a forecast horizon of approximately 70 cycles, and all four models report between 5.9 and 7.5 MAPE, several times their error on Li-ion or Na-ion. The architectural properties that prove advantageous on the larger domains thus do not transfer to this setting.

\begin{figure*}[t]
  \centering
  \includegraphics[width=1.4\columnwidth]{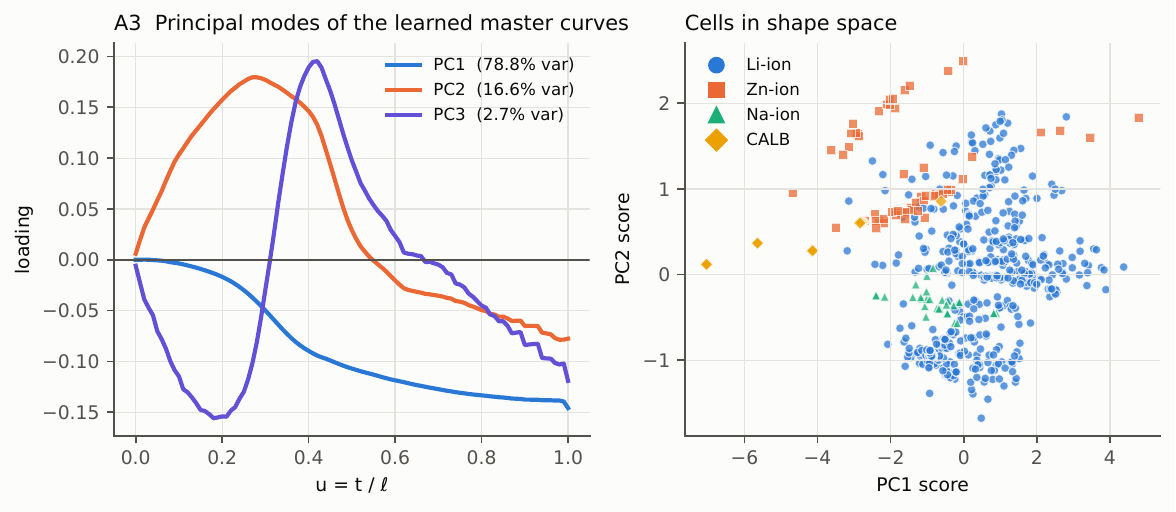}
  \caption{Principal components of the predicted master curves of all 656 test cells. Left: loadings of the first three components, explaining 78.8\%, 16.6\% and 2.7\% of the variance; the first sets the total fade and the second the timing of the descent. Right: cells in the plane of the first two components, by domain.}
  \label{fig:pca}
\end{figure*}

\begin{figure*}[t]
  \centering
  \includegraphics[width=\textwidth]{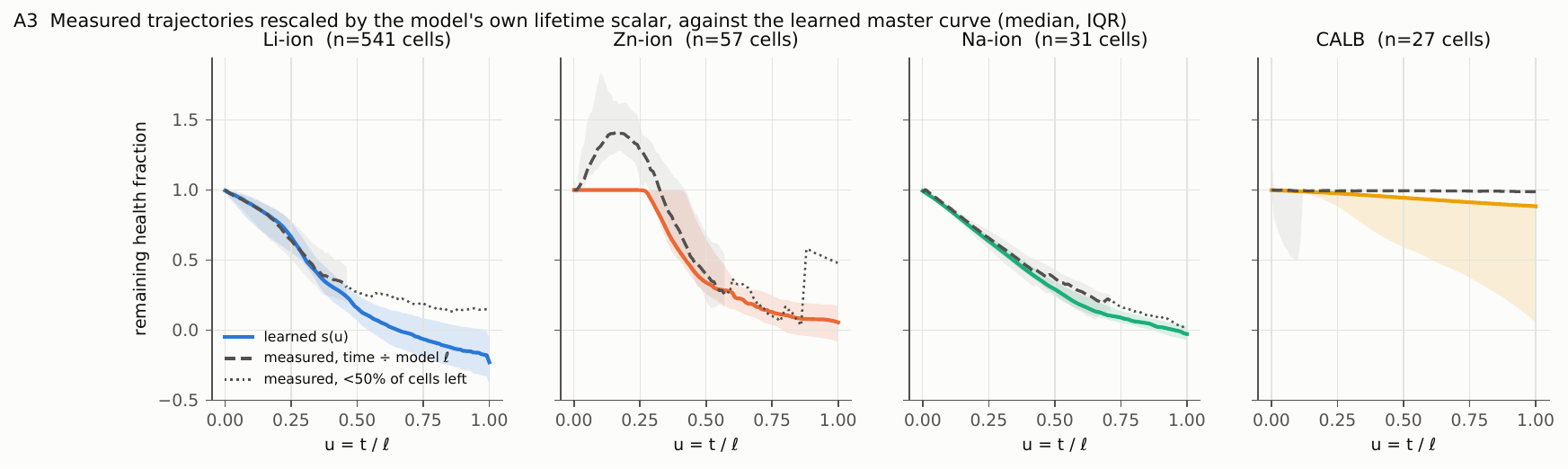}
  \caption{Time lifetime superposition on the test cells of all 22 folds. Measured trajectories, normalised from the initial level (1) to end of life (0), plotted against $u=t/\ell$ using each cell's predicted $\ell$ (black), with the predicted master curve $s(u)$ (colour); medians and interquartile ranges. Dotted where fewer than half of the cells retain data. On Zn-ion, cells gain capacity before fading, which a non increasing master curve cannot represent.}
  \label{fig:collapse}
\end{figure*}

\subsection{Ablation study}
Table~\ref{tab:ablation} varies the side inputs supplied to DecayFormer while holding the architecture, the folds and the protocol fixed. The two configurations differ in their inputs alone; the SOC branch described in the Methodology is present in both. Withholding the measured side inputs, namely the SOC curve channel and the per cycle coulombic and energy efficiencies, degrades Li-ion by 0.143 MAPE and forfeits all three folds, which establishes a material contribution on that domain. On the remaining three domains no contribution is detectable: the two configurations differ by 0.003 on Na-ion and by 0.003 on Zn-ion, in both cases far within the fold dispersion, and the configuration without these inputs is ahead on CALB. On three of the four domains the comparison presented in Table~\ref{tab:final_result} therefore rests upon the architecture rather than upon the side inputs. Supplying BatteryMFormer's Qwen3 embedding of the test protocol text degrades Li-ion, Na-ion and Zn-ion, and improves only CALB, whose fold dispersion is approximately $\pm2.0$ and upon which no claim is advanced. The embedding therefore confers no measurable benefit on this benchmark. BatteryMFormer incurs this dependency whereas DecayFormer does not, requiring neither test protocol metadata nor a language model at inference.

\subsection{Interpreting the master curve head}
Figures~\ref{fig:oracle}, \ref{fig:collapse} and \ref{fig:pca} examine the two outputs of the master curve head on the test cells of all 22 folds.

Correcting the lifetime scalar alone, with the predicted master curve left unchanged, reduces the error on every domain, but by very different proportions (Figure~\ref{fig:oracle}, left). On Li-ion, Na-ion and CALB it removes 56, 62 and 65\% of the error, lowering the MAPE from 2.31 to 1.02, from 0.94 to 0.30 and from 2.81 to 0.54. It also contracts the dispersion across folds, most visibly on CALB, whose standard deviation falls from the largest of any domain to one comparable with that of Li-ion. On these three domains the predicted shape is therefore close to adequate, and most of the remaining error arises from the rate at which a cell is assumed to traverse it. On Zn-ion the same correction removes only 22\%, from 6.76 to 5.30, so that the error remaining under an ideal lifetime still exceeds the full error of the model on every other domain. The Zn-ion deficit therefore lies in the shape of the forecast rather than in its time scale.

The cycle at which each forecast crosses the end-of-life threshold follows the cycle at which the cell did (Figure~\ref{fig:oracle}, right), over more than an order of magnitude on Li-ion, from approximately 100 to 5000 cycles. The points nonetheless lie predominantly above the line of equality: the forecasts reach end of life later than the cells. The displacement is largest for short-lived cells. Li-ion cells that reached end of life within approximately 300 cycles are forecast to do so at approximately 400 to 1000 cycles, whereas the longest-lived cells lie close to the diagonal. The Na-ion and Zn-ion cells occupy the range of approximately 100 to 300 cycles, with a few Zn-ion forecasts crossing near 1000 cycles. This systematic late crossing is consistent with the lifetime scalar accounting for most of the error on Li-ion and Na-ion.

Figure~\ref{fig:collapse} compares the measured trajectories, rescaled by the predicted lifetime, with the predicted master curves. On Li-ion the two medians coincide over the portion of life that most cells record, from $u=0$ to approximately 0.45, and their interquartile ranges overlap throughout. This includes a steeper decline between approximately $u=0.2$ and $0.4$, after which the fade eases, a feature that both curves reproduce. Beyond this point fewer than half of the cells retain data, and the measured median, carried only by the slower cells, supports no comparison. On Zn-ion the measured trajectories first rise, from the initial level to approximately 1.4 times it by $u\approx0.15$, with the upper quartile reaching approximately 1.8, and only then decline. The predicted curve remains at the initial level until $u\approx0.27$, then falls steeply, and meets the measured median near $u\approx0.5$, after which the two agree. The model thus reproduces the timing and steepness of the Zn-ion descent but not the preceding rise, which a non increasing curve cannot follow; over the first part of life it therefore underestimates the state of health, which is the shape error isolated in Figure~\ref{fig:oracle}. On Na-ion the two medians lie almost on top of one another along a nearly linear decline, with narrow interquartile ranges. On CALB the measured trajectories remain close to the initial level across the whole axis, whereas the predicted curves decline gradually to approximately 0.88 at $u=1$, with an interquartile range extending far below the measured one; the model anticipates more fade than these trajectories display.

The predicted master curves vary along few directions (Figure~\ref{fig:pca}). The first two principal components \cite{Pearson,JolliffeCadima} explain 78.8 and 16.6\% of their variance, and the third only 2.7\%. The first loading is close to zero up to $u\approx0.1$ and becomes increasingly negative toward $u=1$; this component changes how far the curve has fallen in late life while leaving early life almost unchanged, and therefore encodes total fade. The second loading is positive over the first half of life, peaking near $u\approx0.28$, and negative beyond $u\approx0.55$; it raises the curve early and lowers it late, and therefore encodes the timing of the descent. The third is a small local correction between approximately $u=0.2$ and $0.45$. In the plane of the first two components the domains occupy distinct regions. CALB cells have the lowest first-component scores, corresponding to the least total fade, in agreement with their nearly flat curves in Figure~\ref{fig:collapse}. Zn-ion cells have the highest second-component scores, corresponding to a delayed descent, the flat-then-steep form seen in their predicted curves. Na-ion cells form a compact group near the origin with slightly negative second-component scores, consistent with a steady fade from the outset. Li-ion cells span the widest range of both components and divide along the second into two groups, centred near $-1$ and between 0 and 1.5. The head thus assigns each chemistry its own region of a two dimensional shape space, and within Li-ion it separates cells whose fade sets in early from those whose fade is delayed.

\section{Conclusion}
This work introduced DecayFormer, a compact architecture that pairs a selective recurrence over cycle tokens with an attention queried from the recurrent state, and predicts a trajectory as a lifetime scalar together with a monotonically non-increasing master curve of remaining health. Across all 22 aging condition exclusive folds of the BatteryLife benchmark, evaluated within BatteryMFormer's released pipeline, it attains the lowest test MAPE of the four models considered on Li-ion and Na-ion, which comprise 15 of the 22 folds, using 0.79M parameters against 12.9 to 14.9M for BatteryMFormer. On Li-ion it is the best of the four models on all three folds, at 2.252, and on Na-ion it is the best on 8 of 12 folds, at 0.950. These results are obtained with a single configuration on every fold, without per fold hyperparameter search, test protocol metadata or a language model. The ablation indicates that the comparison rests on the architecture rather than on its inputs: the measured side inputs contribute materially only on Li-ion, by 0.143 MAPE, and the language model embedding of the test protocol confers no measurable benefit on any domain on which a claim is advanced.

DecayFormer is outperformed on Zn-ion, at 7.185 against 5.907 for BatteryMFormer, and no ordering is supported on CALB, whose public records terminate at 99 to 100 cycles. The analysis of the master curve head clarifies both its strength and the origin of the Zn-ion deficit. Rescaling cycle index by the predicted lifetime collapses the measured trajectories onto the learned master curves, which supports the time lifetime superposition on which the head is built, and on the larger domains most of the remaining error arises from the lifetime scalar rather than from the shape. On Zn-ion the error lies instead in the shape: the cells gain capacity before they fade, and a master curve constrained to be non increasing cannot represent that rise. Two directions follow. Relaxing the monotonicity constraint over the early part of life would allow the head to represent this behaviour without forfeiting the robustness that the constraint provides elsewhere, and improving the estimate of the lifetime scalar is the most direct route to further gains on the remaining domains. Evaluating the architecture on the BatteryML benchmark \cite{BatteryML}, and against PhyMamba \cite{phyMamba}, would further establish its generality across datasets and model families.

\end{document}